\documentclass[runningheads]{llncs}
\usepackage{graphicx}

\usepackage{tikz}
\usepackage{comment}
\usepackage{color}

\usepackage{booktabs}

\usepackage{epsfig}
\usepackage{graphicx}
\usepackage{grffile}
\usepackage{animate}
\usepackage{amsmath}
\usepackage{amssymb}
\usepackage{colortbl}
\usepackage{helvet}
\usepackage{courier}
\usepackage[export]{adjustbox}
\usepackage{bm}
\usepackage{multirow}
\usepackage{caption}
\usepackage{booktabs}
\usepackage{color}
\usepackage{dsfont}
\usepackage{booktabs}
\usepackage{xcolor}
\usepackage{algorithm}
\usepackage{algorithmic}
\usepackage[normalem]{ulem}
\usepackage{diagbox}
\usepackage[hoptionsi]{overpic} 
\newcommand{\tabincell}[2]{\begin{tabular}{@{}#1@{}}#2\end{tabular}}

\newcommand\ies{\textit{i.e.}}
\newcommand\egs{\textit{e.g.}}

\newcommand\figcaption{\def\@captype{figure}\caption}
\newcommand\tabcaption{\def\@captype{table}\caption}
\definecolor{citecolor}{HTML}{0071bc}
\usepackage[capitalize]{cleveref}
\usepackage[accsupp]{axessibility}  

\begin{document}
\pagestyle{headings}
\mainmatter

\title{Thunder: Thumbnail based Fast Lightweight Image Denoising Network} 

\titlerunning{Thunder}
%
\author{Yifeng Zhou\inst{1} \and
Xing Xu\inst{1} \and
Shuaicheng Liu\inst{1}\and Guoqing Wang\inst{1} \and Huimin Lu\inst{2} \and Heng Tao Shen\inst{1} }
\authorrunning{Zhou. Author et al.}
%
\institute{Center for Future Multimedia and School of Computer Science and Engineering, \\
University of Electronic Science and Technology of China, China \and
Department of Mechanical and Control Engineering, Kyushu Institute of Technology, Japan
}

\maketitle

\begin{abstract}
  To achieve promising results on removing noise from real-world images, most of existing denoising networks are formulated with complex network structure, making them impractical for deployment. Some attempts focused on reducing the number of filters and feature channels but suffered from large performance loss, and a more practical and lightweight denoising network with fast inference speed is of high demand.
  To this end, a \textbf{Thu}mb\textbf{n}ail based \textbf{D}\textbf{e}noising Netwo\textbf{r}k dubbed Thunder, is proposed and implemented as a lightweight structure for fast restoration without comprising the denoising capabilities. Specifically, the Thunder model contains two newly-established modules:
  (1) a wavelet-based Thumbnail Subspace Encoder (TSE) which can leverage sub-bands correlation to provide an approximate thumbnail based on the low-frequent feature; (2) a Subspace Projection based Refine Module (SPR) which can restore the details for thumbnail progressively based on the subspace projection approach.
  Extensive experiments have been carried out on two real-world denoising benchmarks, demonstrating that the proposed Thunder outperforms the existing lightweight models and achieves competitive performance on PSNR and SSIM when compared with the complex designs. 
\end{abstract}

\section{Introduction}
\label{sec:intro}

Image denoising aims to restore the high-quality image from the noisy observation. 
Recently, with the advance of deep neural networks, various models \cite{NBNet,MIRNet,mehri2021mprnet,vdn,danet} have achieved impressive results on real-world noisy datasets. 
Despite the emergence of many effective methods, most of them consists of complex modules with large amount of parameters and call for huge computational cost in practice.
Denoising is served as the pre-processing method for the camera pipeline, especially for smartphones.
However, the existing heavy-loaded models can hardly be deployed on mobile devices, which arouses the demand of developing efficient and lightweight denoising methods.

\begin{figure}[t]
    \centering
    \scriptsize
    \begin{minipage}[b]{0.55\textwidth}
        \centering
        \includegraphics[width=\linewidth]{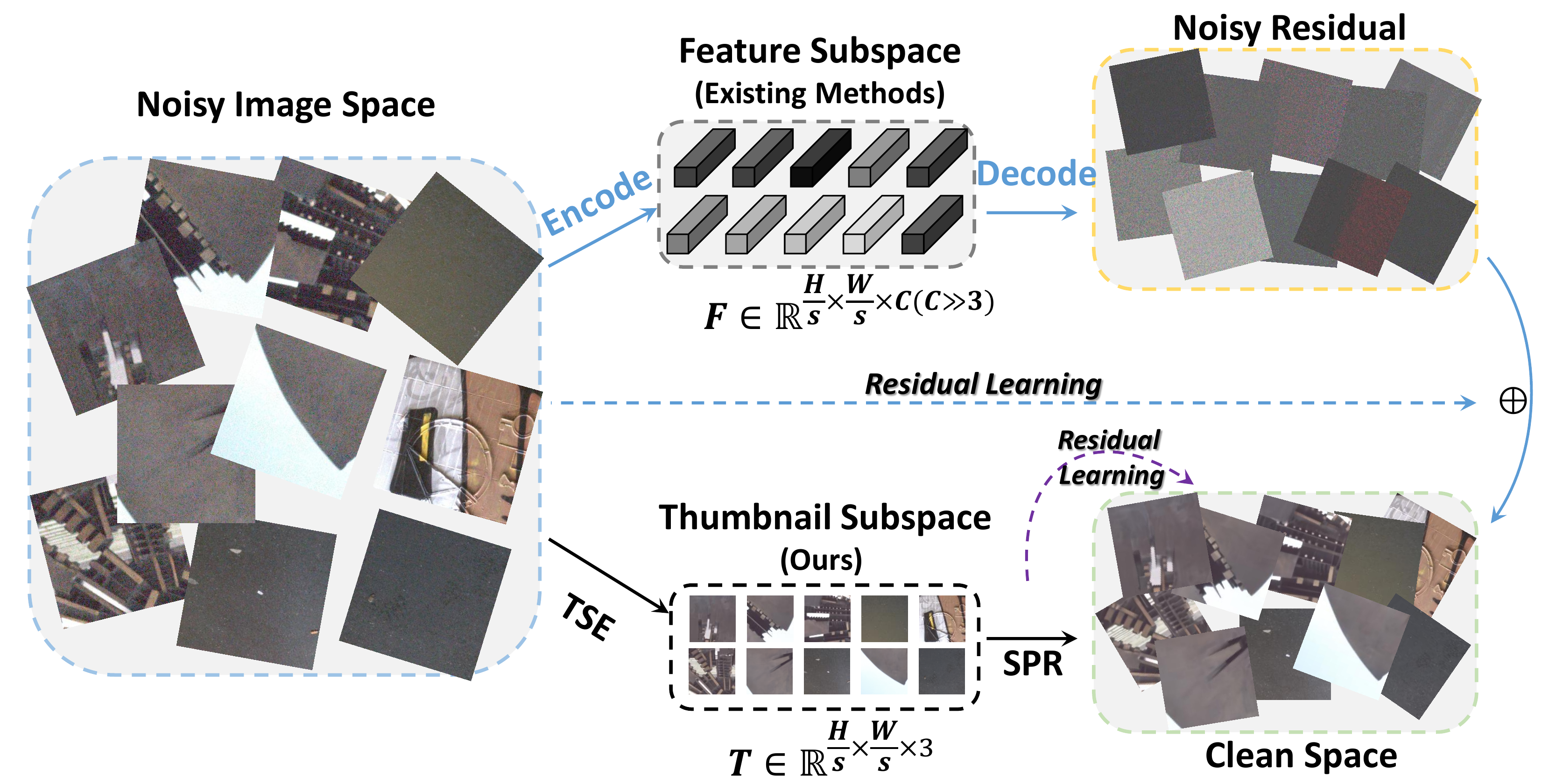}
    \end{minipage}
    \hfill
    \begin{minipage}[b]{0.42\textwidth}
        \centering
        \includegraphics[width=\linewidth]{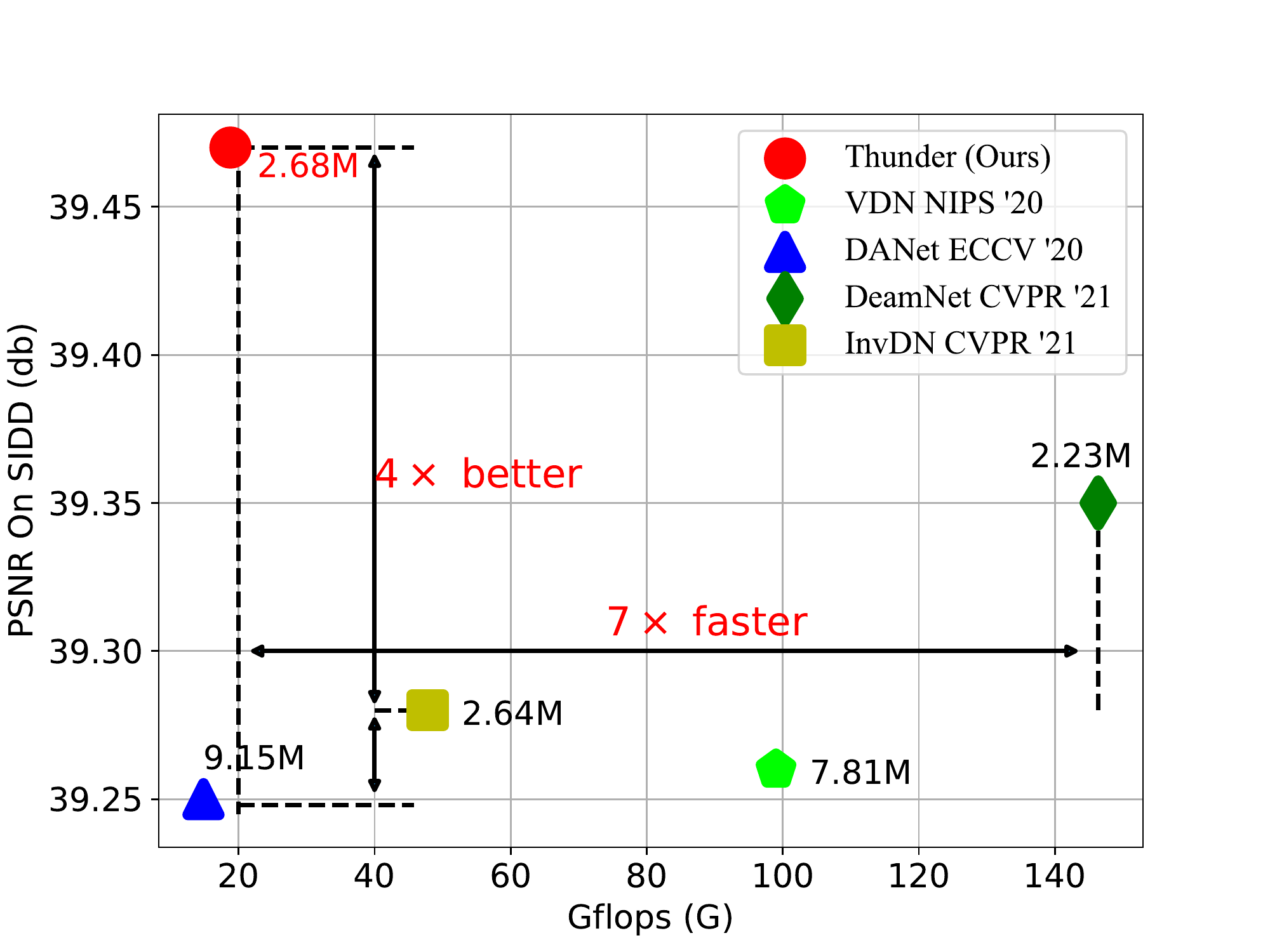}
    \end{minipage}
    \caption{
        \textbf{Analysis of parameters}.
        (Left) Existing UNet-based methods (Blue lines) encode the noisy image to feature subspace and decode this to conduct the end-to-end residual learning, which demands a huge amount of channel numbers ($C\gg3$) to describe the noisy-aware and signal-aware features. Thunder leverages TSE and SPR to perform the denoising in two stages to reduce the complexity.
        (Right) Comparison of the proposed Thunder method and the latest lightweight approaches in terms of PSNR, GFlops and Parameters. The numbers near each point denotes the parameter's amount of correspond model.
    }
    \label{fig:pipeline_and_cmp}
\end{figure}

	

In order to accelerate the denoising process meanwhile maintain the model capability, two promising designs have been proposed including (1) The residual-based learning methods \cite{ffdnet,dncnn,danet,NBNet,MIRNet}. Instead of estimating the clean image $X$ from a noisy input $Y$, they learn the noisy residual $N = Y - X$.
Such designs can guarantee an easier optimization process.
(2) The UNet-like framework \cite{unet} which can save the computational costs meanwhile enlarging the receptive filed.
Existing methods leverage these two aspects, specifically, as demonstrated by the blue lines in Fig. \ref{fig:pipeline_and_cmp}. They first project the noisy image into feature subspace: $\bm{F} \in \mathbb{R}^{\frac{H}{s}\times \frac{W}{s}\times C}$; $C\gg3$, where $H$ and $W$ denote the size of the noisy image, $C$ denotes the channel numbers for the feature and $s$ denotes the down-sample scale. 
Then, the restored images can be generated from the projected features.
Although these methods can achieve satisfying performance, their complexity increases sharply.

Some observations encourage us to polish the above two designs to get a more effective approach.
For the residual learning approach, existing methods proposed to use the end-to-end residual learning mechanism. When the model estimates the noise residual, it implicitly estimates the signal, which doubles the computational burden.
To decrease the model complexity, we first decompose the low and high frequent features, then only the low frequent feature are leveraged to generate the thumbnail. Finally, instead of using the end-to-end residual learning, we conduct it in the signal reconstruction stage.
This can alleviate the model from struggle concurrent learning of the noise distribution and the signal details.
In principle, considering, due to the spatial-correlation of signal, the thumbnail is easy to obtained and more likely to reconstruct the signal, while the denoising procedure can be the by-product of the down-sample operation.

Moreover, the previous end-to-end residual includes the noise reduction and the signal restoration. Different from this, we depart them into two stages. 
As shown in Fig. \ref{fig:pipeline_and_cmp}, the noise reduction is achieved by Thumbnail Subspace Encoder (TSE), which produces the small thumbnail with less noise; the signal restoration is achieved by Subspace Projection based Refine Module (SPR) which learns the signal residual to refine the thumbnail.
For the UNet-like framework, the existing methods suffer from huge computational cost because convolutional layers project the feature by exponentially increasing channel numbers ($3 \rightarrow 32 \dots \rightarrow 1024$). 
To achieve fast denoising with a lightweight network, we compress the feature subspace to a small RGB subspace: $\bm{T} \in \mathbb{R}^{\frac{H}{s}\times \frac{W}{s}\times 3}$, therefore, we can alleviate a lot of computational costs by reducing the channel numbers for all the convolutional layers.
As discussed in \cite{Noise_level_estimation,Noise_level_estimation2}, the noise is uniformly distributed across the ambient space, while the noiseless images (signals) lie in a low-dimensional subspace because of the spatial correlation.
Based on this assumption, we leverage the project operation to restore the details for the thumbnail.

Specifically, we propose a lightweight denoising network termed \textbf{Thu}mb\textbf{n}ail based \textbf{D}\textbf{e}noising Netwo\textbf{r}k (\textit{Thunder}), which leverages the RGB thumbnail $\bm{T}$ instead of the feature subspace $\bm{F}$ to accelerate the denoising process.
As shown in Fig. \ref{fig:pipeline_and_cmp}, the proposed Thunder reconstructs the signal progressively by employing the novel TSE which compresses the complexity, and SPR which learns clean signal subspace projection.
The proposed Thunder is able to achieve promising denoising results comparable to the complex models: \textbf{39.47dB} on SIDD test dataset with much less parameters \textbf{2.68M} and \textbf{18.81} GFlops, shown in Fig. \ref{fig:pipeline_and_cmp}.

The main contribution of this paper are as follows:
\begin{itemize}
	\item We introduce the Thunder network which accelerates the denoising process based on the thumbnail subspace. Instead of learning  end-to-end residual, we propose to divide the denoising process into two steps, and such formulation can significantly reduce the complexity without compromising its denoising capabilities.
	
	\item We put forward the Wavelet-based Thumbnail Subspace Encoder (TSE) which can encode the noisy image to thumbnail space in terms of the sub-bands correlation and meanwhile reducing the noise.
	
	\item We propose the Subspace Projection based Refine Module (SPR) which can progressively restore the thumbnail based on the projection matrix.
\end{itemize}

\section{Related Work}
\label{sec:rel}
\subsection{PSNR-aware Denoising Methods}
The target of the denoising task is to achieve higher PSNR. 
In the early time, the non-local patch similarity is widely used in the traditional  \cite{nonLocalMean,BM3D} methods, and more patches are selected for feature learning and producing better results.
Also, some Gaussian Mixture models \cite{otherTranditional1Gau,otherTranditional2Gau} are proposed to solve the Gaussian additive noise.

Later, with the development of the deep neural network, the CNN-based denoising methods show the dominant advantages over the traditional methods.
DnCNN \cite{dncnn} is the first to apply the CNN to image denoising , and impressive results can be obtained by stacking some convolutional blocks for learning the noise residual.
CBDNet \cite{CBDNet} leverages the noise level estimation sub-net and the multi-scale feature to denoise the real-world noisy images.
MIRNet \cite{MIRNet} enhances the correlation between different scale branches and applies the dual attention block.
NBNet \cite{NBNet} proposes a subspace projection method to preserve the signal details while reducing the noise.
Some patch-based multi-stage manners \cite{mehri2021mprnet,MultiPatchDeblur2019,MultiPatchDblur2020} are used to restore the image progressively.
Similarly, some recursively branched-based methods are applied to refine the outputs, such as RBDN \cite{RBDN}.
Since the noiseless image is difficult to obtain, there emerge many transfer learning methods from synthetic to real denoising \cite{denomoraliseDenoise,denomoraliseDenoise2,zhou2020awgn}.
It takes more channel numbers (filters numbers) $\bm{F} \in \mathbb{R}^{\frac{H}{s}\times \frac{W}{s}\times C}; C\gg3$ to represent the noisy feature when applying the noisy residual-based approach. 
The proposed Thunder leverages the thumbnail to significantly reduce the complexity of existing models and well preserve the capability of denoising.

\subsection{Efficiency-aware Denoising Methods}
While pushing the PSNR higher, the practicality of the denoising methods has drawn more and more attention.
As the fast version of DnCNN the FFDNet \cite{ffdnet} down-sample the input to reduce the computational cost.
DANet and C2N \cite{danet,c2n} propose the GAN-based manner to approximate the noisy distribution while the generator can boost the performance of the lightweight denoiser.
VDN \cite{vdn} optimizes the variational lower bound to achieve the fast blind real-world denoising.
MWCNN \cite{MWCNN} replaces the pooling layer with the wavelet transform, for better restoration in the frequency domain. 
Other approach \cite{wang2020practical} leverages the depth-wise convolutional layer to accelerate the network. 
Besides, some methods \cite{deamnet} leverage adaptive consistency prior to obtain the comparable performance with fewer parameters.
To extend the invertibility of the network, \cite{invdn} leverage the invertible network designs to reduce the parameter budget.
However, the forward and backward computational cost is the same as the uninvertible model.
All the above-mentioned lightweight designs suffer from being unable to produce promising quantitative results due to the incapability of separating noise and signal during learning.
Our method can decompose the high-frequent (noise, partial signal) and low-frequent (partial signal) in TSE, then restore the signal and reduce the noise progressively in SPR.

\section{Our Thunder Model}

\begin{figure*}[t]
	\centering
	\includegraphics[width=\textwidth]{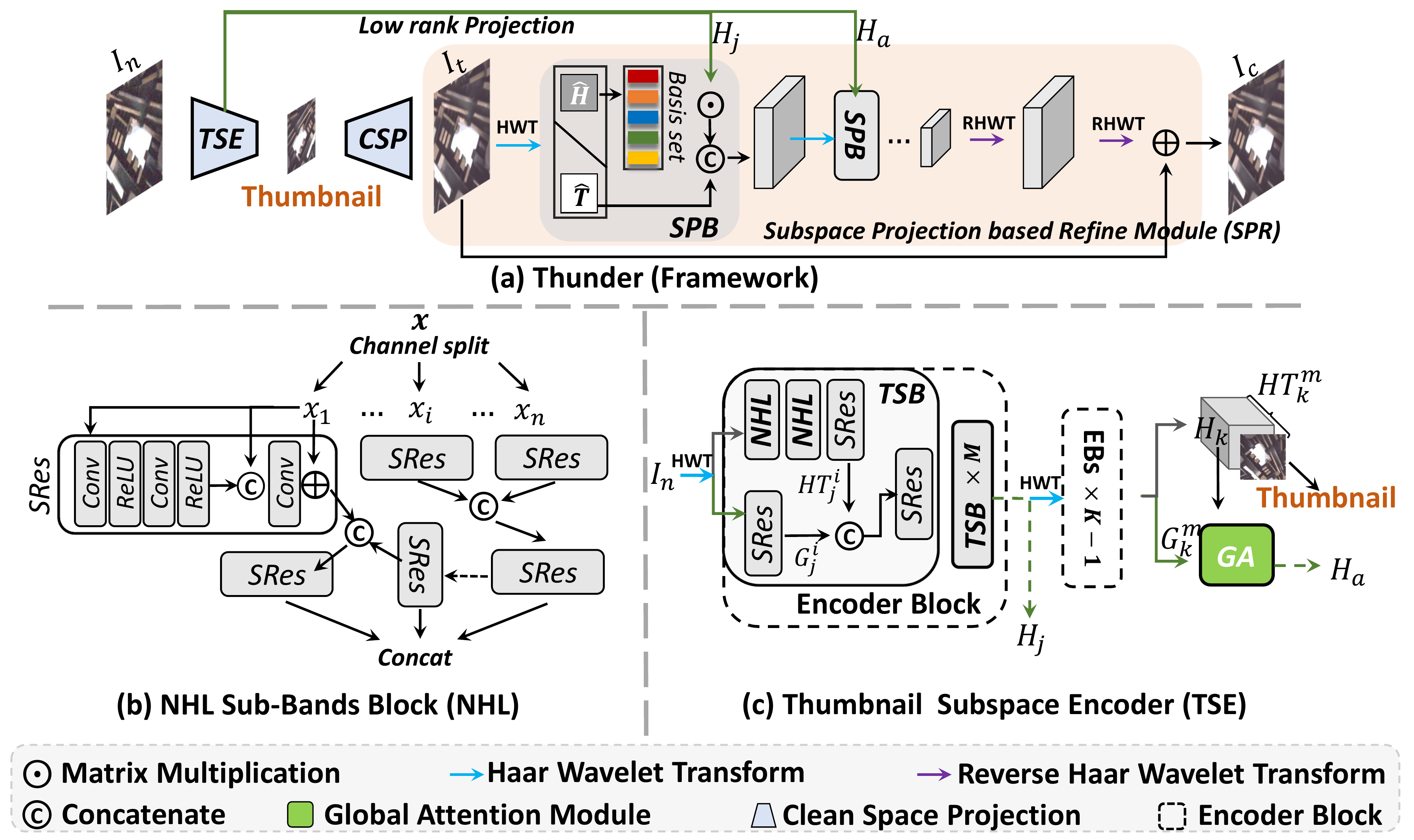}
	\caption{Illustration of (a) The framework of Thunder, which contains the four key components: (b) NHL Sub-Bands Block (NHL); (c) Thumbnail Subspace Encoder (TSE). The GA and CSP are two small sub-nets composed of several $1 \times 1$ convolutional layers, more details are provided in the supplementary materials.
	}
	\label{fig:framework}
	\vspace{-4mm}
\end{figure*}
\label{sec:method}
\subsection{Overall Framework}
As shown in Fig. \ref{fig:pipeline_and_cmp}, the thumbnail-based denoising process is divided into two steps: (1) the thumbnail subspace projection; (2) the thumbnail-based signal reconstruction.
The framework is shown in Fig. \ref{fig:framework} (a).
Given a noisy image $\bm{I_n}$, we first feed it into the Thumbnail Subspace Encoder (TSE), which is composed of \textbf{K} Encoder Blocks (EB).
Each EB contains \textbf{M} Two Stream Blocks (TSB).
Before passing to EB, the Haar wavelet transform (HWT) is used to extract the low and high frequent features as well as reduce the spatial scale. 

After passing through \textbf{K} EBs, the first three channels of each EB output are defined as thumbnail feature $\bm{T_j} (1<j<K)$, and other channels are defined as $\bm{H_j}$ for restoring the signal. 
We term it `Thumbnail' instead of `low-frequency component' because the former is generated from the latter with learned parameters.
Next, the final EB's output high-frequent feature $\bm{H_k}$ is enhanced by the Global Attention Module (GA) to get  $\bm{H_a}$.
To match the spatial scale and connect the thumbnail subspace and corresponded clean image (Ground-Truth) subspace, the Clean Space Projection sub-net (CSP) is used to project the final EB's output thumbnail $\bm{T_k}$ to $\bm{I_t}$.
Finally, $\bm{I_t}$ and all the high-frequent features of each EB ($\bm{H_j}$ and $\bm{H_a}$) are fed into the Subspace Projection based Refine Module (SPR) to get the clean image $\bm{I_c}$.

\subsection{Thumbnail Subspace Encoder}
\subsubsection{Encoder Block}
In the beginning, the HWT is used to decompose the $\bm{I_n} \in \mathbb{R}^{ H \times W\times 3}$ from the frequency domain to obtain $\bm{HT_j^i} \in \mathbb{R}^{\frac{H}{k_j}\times \frac{W}{k_j}\times 3C_j}$, where the $i,j$ denotes the $i^{th}$ TSB in $j^{th}$ EB, $C_j$ and $K_j$ is 4 and 2 to the $j$ power, respectively.
We regroup $\bm{HT_j^i}$ into the low-signal feature $\bm{T_j^i} \in \mathbb{R}^{\frac{H}{k_j}\times \frac{W}{k_j}\times 3}$, high-signal feature $\bm{S_j^i}\in \mathbb{R}^{\frac{H}{k_j}\times \frac{W}{k_j}\times 3}$, and noise $\bm{N_j^i} \in \mathbb{R}^{\frac{H}{k_j}\times \frac{W}{k_j}\times 3C_j-6}$, respectively.
This feature is fed into the Two Stream Block (TSB) to extract the inter-correlation and the global information between sub-bands, shown in Fig.\ref{fig:framework} (c) gray and green line.
Note that, the output $\bm{N_j^m}$ and $\bm{S_j^m}$ is combined to get the high-frequent feature: $\bm{H_j}$, then skip-connect to SPR to restore the details of latent clean images.
After \textbf{K} EBs the first three channels of the final EB output $\bm{T_k}$ are defined as thumbnail $\bm{T}$ and fed into CSP to get $\bm{I_t}$. 
The other high-frequency channels $\bm{H_k}$ are sent to the Global Attention Module.

\noindent\textbf{Two Stream Block.}
The $\bm{HT_j^i}$ is processed by two different branches within TSB.
Different frequent features are extracted by different channel at the same spatial position, and each channel represents a sub-band in frequency domain. 
As shown in Fig. \ref{fig:framework} (c), the first branch of TSB is designed to divide the frequent feature into different groups to better exploit the inner sub-bands correlation using NHLs. 
The second branch (green line) is to extract the global sub-bands features $\bm{G_j^i}$. 
Instead of only considering the sub-bands independently, this branch takes the whole sub-bands into consideration.

\noindent\textbf{NHL Sub-Bands Block.}
Due to that the Haar Transform is carried out to produced coarse decomposition in the frequency domain, the features passed by the low-pass filters also contain the high-frequent details.
To boost the propagation of features forward to their correct bands, we propose the NHL block. 
Different from previous wavelet-based denoising methods \cite{MWCNN,invdn,manet}, and inspired by \cite{subbands}, NHL leverages the correlation between sub-bands.
As shown in Fig. \ref{fig:framework} (b), the decomposed features $\bm{T_j^i}$, $\bm{S_j^i}$ and $\bm{N_j^i}$, are firstly processed by three Self-Residual Blocks (SRes) $\Omega(\cdot)$ independently.
Then, the inter-correlation between each sub-band, which is denoted as $\Phi(\cdot)$, can be computed by the self-residual recursively:
\begin{equation}
\label{NHL2}
\Phi(\bm{A}, \bm{B}) = 	\Omega_{3}(Cat(\Omega_{1}(\bm{A}), \Omega_{2}(\bm{B}))),
\end{equation}
where $\bm{A}$ and $\bm{B}$ can be any sub-bands features ($\bm{N_j^i}$, $\bm{S_j^i}$ and $\bm{T_j^i}$).
Based on Eq. \ref{NHL2}, we leverage the correlation between $\bm{N_j^i}$ and $\bm{S_j^i}$ to obtain the new signal features.
Then, we use the enhanced $\bm{S_j^{'i}}$ and $\bm{T_j^i}$ to compensate the low-signal features in the same way.
In Eq.\ref{NHL2}, the divided features (T, S, N) are processed independently, which can be viewed as first-order sub-bands correlation modeled by $\Omega(\cdot)$. 
Then, $\Phi(\cdot)$ is used to get the second-order feature. 
Finally, the global branch can be regarded as the third-order (global) feature to conduct the information propagation.
The self-residual operation $\Omega(\cdot)$ is defined as:
\begin{equation}
\label{NHL}
\Omega(\bm{I}) = R(\bm{I}) + \bm{I},
\end{equation}
where the $R(\cdot)$ can be any network. (\egs, the residual operation in our implementation), and $\bm{I}$ denotes the input.

\subsubsection{Global Attention and Clean Space Projection}
After obtaining $\bm{G_k^m}$, $\bm{H_k^m}$ and $\bm{T}$ at final EB, $\bm{H_k^m}$ and $\bm{G_k^m}$ are passed through the GA to get attentive high-frequent features.
The GA subnet is composed of a stack of convolutional layers with the size of kernels set as $1 \times 1$.
This convolution can be viewed as fully connected layer for all the frequent channels at each spatial position. 
Then the attentive high-frequent $\bm{H_a}$ feature is given by:
\begin{equation}
\label{GA}
\bm{H_a} = 	GA(\bm{G_k^m}) * \bm{H_k^m} + \bm{H_k^m}.
\end{equation}
Due to the difference between the scale of $\bm{T}$ and the original scale, the CSP subnet is applied to project $\bm{T}$ to the same spatial-scale clean subspace.
The $\bm{I_t} \in \mathbb{R}^{{H}\times {W}\times 3} $ can be obtained by $CSP(\bm{T})$.
The design of CSP is inspired by some super-resolution framework \cite{srlut}, and several convolutional layers followed by a pixel shuffle layer are utilized to produce the $\bm{I_t}$.
Due to the space limitation, the details for GA and CSP are provided in the supplementary material.

\subsection{Subspace Projection Based Refine Module}
For the signal reconstruction stage, we apply the SPR to refine $\bm{I_t}$, as shown in Fig. \ref{fig:framework} (a).
It is worth to be noted that, Thunder also performs the residual learning, however the residual is not for noise but signal.
The SPR is also designed in a UNet manner, and the final output is added back to $\bm{I_t}$.
The proposed SPR contains K Space Projection Blocks (SPB). 
Similar to TSE, HWT is also exploit here to separate high $\bm{\hat{H}} \in \mathbb{R}^{\frac{H}{k_j}\times \frac{W}{k_j}\times (4C_j-3)}$ and low $\bm{\hat{T}} \in \mathbb{R}^{\frac{H}{k_j}\times \frac{W}{k_j}\times 3}$ frequent features of $\bm{I_t}$.
The target of SPR is to refine the $\bm{\hat{H}}$. 

To implement this, we leverage the high-frequent features $\bm{H_j}$ which extract by TSEs (Noted that $\bm{H_j}$ includes $\bm{H_a}$).
However,  $\bm{H_j}$ contains both the high-frequent signal feature (image details) and the noise.
Therefore, the subspace projection method is utilized to separate the signal information and the noise.
Different from \cite{NBNet} which projects the noisy feature to signal subspace and produce the noisy residual, we conduct the projection within high-frequent signal space and use the projected feature to refine the signal.
In this way, the inconsistency between inner features (\ies, the signla related ones) and output residuals (\ies, the noisy related ones) can be avoided to achieve better performance.
Specifically, SPB aims to project $\bm{H_j}$ to into the signal subspace. 
Firstly, the subspace basis vector $v_i$ is estimated based on $\bm{H_j}$ and $\bm{\hat{H}}$ to compose the Basis set $\bm{V} =[v_1, v_2 \dots, v_q ], \in \mathbb{R}^{HW \times Q}$.
Then, the $\bm{H_j}$ is transformed into the signal subspace by orthogonal linear projection:
\begin{equation}
\label{SP}
\tilde{\bm{H}} = \bm{V}(\bm{V}^{T}\bm{V})^{-1}\bm{V}^{T}\bm{H_j},
\end{equation}
where $\tilde{H}$ is the projected high-frequent feature.
After projection, the low dimensional signal is more likely to be preserved, while the noise will be reduced.
Next, we concatenate projected high-frequent feature with the original low-frequent feature and fed it to NHLs to restore the sub-bands correlation.
Finally, the clean image $\bm{I_c} \in \mathbb{R}^{H \times W\times 3}$ can be computed by several Reversed HWTs and NHLs.

\subsection{Objective Function}
To train the proposed Thunder, a hierarchical loss is formulated inlcuding the thumbnail projection loss $\mathcal{L}_t$, and the refinement loss $\mathcal{L}_r$:
\begin{equation}
\label{total}
\mathcal{L} = \mathcal{L}_t + \underbrace{\alpha\mathcal{L}_{1(c)} + \beta\mathcal{L}_{1(t)} + \mathcal{L}_{G} + 	\mathcal{L}_{S}}_ {\mathcal{L}_r}.
\end{equation}
In Eq.\ref{total}, $\mathcal{L}_t$ is the L2 loss which is applied to obtain a clean thumbnail. 
\begin{equation}
\label{L2}
\mathcal{L}_t = \| \bm{X_s} - \bm{T} \|_2,
\end{equation}
where  $\bm{T}$ is the thumbnail and $\bm{X_s}$ is $\frac{1}{s}$ down-sample from clean image $\bm{X}$.
$\mathcal{L}_r$ is composed by some sub-losses.
The first is L1 distance between clean images and the denoising result:
\begin{equation}
\label{L1}
\mathcal{L}_{1(o)} = \| {\bm{I_o}} - \bm{X} \|_1;  o \in \{t,c\},
\end{equation}
where $\bm{I_o}$ represents the coarse clean image $\bm{I_t}$ or the final refined output $\bm{I_c}$, respectively. 
Considering the detail reconstruction process of Thunder, the orientation-aware constraint is also developed and implemented as the gradient loss as follows:
\begin{equation}
\label{Gradient}
\mathcal{L}_{G} = \| \Delta_{x/y}{\bm{I_c}} - \Delta_{x/y}\bm{X} \|_1,
\end{equation}
where $\Delta_{x/y}$ denotes the gradient computation along horizontal or vertical.
To better preserve the structural details of the output image, the SSIM loss is used:
\begin{equation}
\label{SSIM}
\mathcal{L}_{S} = 1- SSIM(\bm{I_c}, \bm{X}),
\end{equation}
the $SSIM(\cdot)$ denotes the SSIM value computed between the input groundtruth and the denoising output.
\begin{table*}[t]
	
	\centering
	
	\resizebox{\linewidth}{!}{
		\begin{tabular}{|c|cc|c|c|c|cc|cc|}
			\hline
			\multirow{2}{*}{Type}&\multicolumn{2}{c|}{Method} & &  &  & \multicolumn{2}{c|}{SIDD \cite{sidd}} & \multicolumn{2}{c|}{DND  \cite{dnd}}\\
			\cline{2-3}
			\cline{7-10}
			&Model & Years & \multirow{-2}{*}{Traing Dataset} & \multirow{-2}{*}{Params  $\downarrow$} & \multirow{-2}{*}{GFlops  $\downarrow$} & PSNR $\uparrow$ & SSIM $\uparrow$ & PSNR $\uparrow$ & SSIM  $\uparrow$\\
			\hline

			\multirow{3}{*}{Heavy-Loaded}&AINDNet~\cite{denomoraliseDenoise2} &2020& SIDD+Gaussian & 13.76 & -- & 39.08 & 0.953 & 39.53 & 0.9561  \\ 
			&MIRNet~\cite{MIRNet} &2020 & SIDD  & 31.78 & 1569.88 & -- & -- & 39.88 & 0.956  \\
			
			&MPRNet ~\cite{mehri2021mprnet}& 2021& SIDD  & 15.74 & 588.14 & -- & -- & {39.82} & 0.954  \\

			&NBNet ~\cite{NBNet} &2021& SIDD  & 13.31 & 88.79  & -- & -- & \textbf{{39.89}} & \textbf{{0.9563}} \\
			\hline
			
			\multirow{10}{*}{Lightweight}&DnCNN ~\cite{dncnn}& 2017 & Gaussian & 0.56 & -- & 23.66 & 0.583 & 32.43 & 0.7900  \\ 
			
			&FFDNet~\cite{ffdnet} &2018 & Gaussian  & 0.48 & -- & -- & -- & 34.40 & 0.8474  \\ 
			
			&CBDNet ~\cite{CBDNet}&2019& RENOIR+Syn.  & 4.34 & 80.76  & 33.28 & 0.868 & 38.06 & 0.9421 \\ 
			
			
			&RIDNet~\cite{RIDNet} &2019& SIDD  & 1.49 & 196.52  & -- & -- & 39.26 & 0.9528 \\ 
			
			&VDN ~\cite{vdn}&2019 & SIDD & 7.81 & 99.00 & 39.26 & 0.955 & 39.38 & 0.9518  \\ 
			&DANet~\cite{danet} &2020& SIDD+RENOIR+Poly+Syn. & 9.15 & 14.85 & 39.25 & 0.916 & 39.47 &	0.9548   \\ 
			
			&DeamNet~\cite{deamnet} &2021& SIDD+RENOIR & 2.23 & 146.36 & 39.35 & 0.955 & \textbf{39.63} & {\textbf{0.955}}  \\ 
			

			&InvDN ~\cite{invdn}& 2021& SIDD & 2.64 & 47.80 & 39.28 & 0.955 & 39.57 & 0.9522  \\ 
			\cline{2-10}
			&Thunder & Ours & SIDD & 2.68 & 18.81 & \textbf{{39.47}} & \textbf{{0.957}} & 39.57 & 0.9526 \\ 
			\hline
		\end{tabular}
	}
	\caption{Comparison of existing denoising methods. `Params' denotes the parameters amount of the corresponding model. `GFlops' presents computational cost for peer $256 \times 256$ image. The best results are emphasized in bold. The model whose parameters amount is larger than 10M is defined as a heavy-loaded model. ``-" denotes not available in the leader board.}
	\label{Tab:DnD_SIDD}
    \vspace*{-4mm}
\end{table*}

\begin{figure*}[!htb]
	\begin{center}
		\small
		\begin{tabular}[t]{c@{ }c@{ }c@{ }c@{ }c@{ }c@{ } c@{ }}
			
			\includegraphics[width=.15\linewidth]{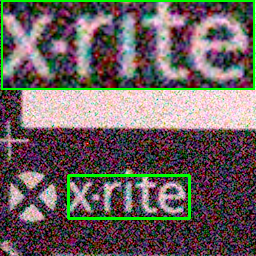}&  
			\includegraphics[width=.15\linewidth]{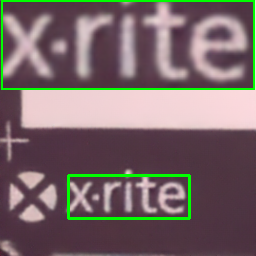}&   
			\includegraphics[width=.15\linewidth]{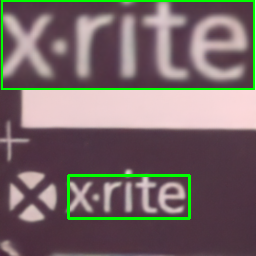}&  
			\includegraphics[width=.15\linewidth]{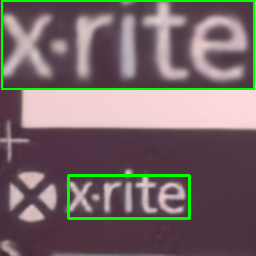}&  
			\includegraphics[width=.15\linewidth]{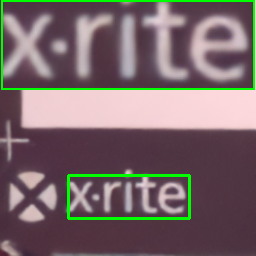}&   
			\includegraphics[width=.15\linewidth]{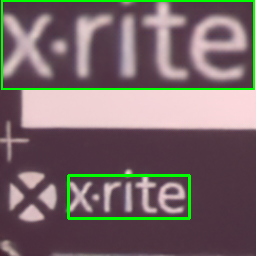}&  	
			\\
			17.61dB &  32.04dB &  32.17dB &  32.14dB &   32.60dB &   \textbf{32.72dB}\\
			Noisy  & DANet  & VDN  & Invdn  &  DeamNet  &  Thunder  \\
			
			\includegraphics[width=.15\linewidth]{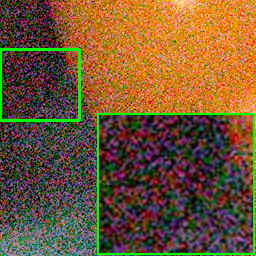}&  
			\includegraphics[width=.15\linewidth]{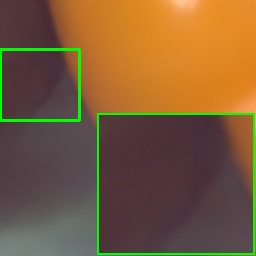}&   
			\includegraphics[width=.15\linewidth]{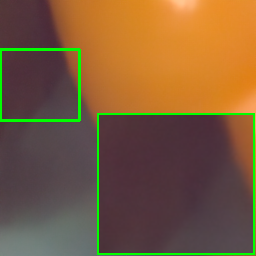}& 
			\includegraphics[width=.15\linewidth]{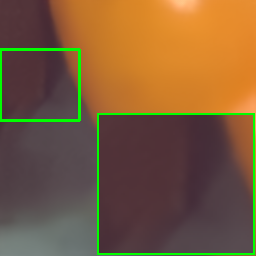}&  			
			\includegraphics[width=.15\linewidth]{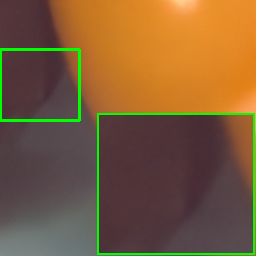}&  
			\includegraphics[width=.15\linewidth]{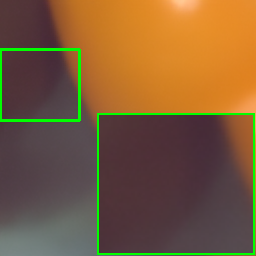}&  
			\\
			\vspace{0.2mm}
			16.18dB & 30.44 dB & 30.35dB &  30.33dB &  30.50dB &   \textbf{30.55dB} \\
			Noisy  & DANet  & VDN  & Invdn  &  DeamNet  &  Thunder  \\
			
			
			
		\end{tabular}
	\end{center}
	\vspace{-4mm}
	\caption{Denoising examples from SIDD. Green boxed regions are zoomed results for detailed observation.}
	\vspace{-4mm}
	\label{fig:sidd_dnd_example}
\end{figure*}

\begin{figure*}[t]
	\begin{center}
		\small
		\scalebox{1}{
			\begin{tabular}[b]{c@{ } c@{ }  c@{ } c@{ } }\hspace{.5mm}
				\multirow{3}{*}{\includegraphics[width=.415\textwidth,valign=t]{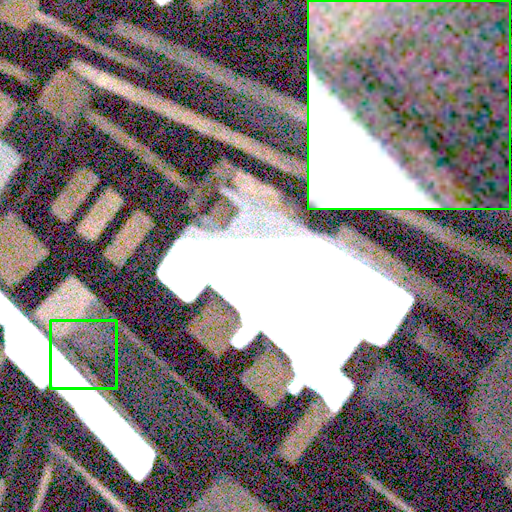}} 
				&  \includegraphics[width=.17\textwidth,valign=t]{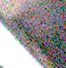} 
				& \includegraphics[width=.17\textwidth,valign=t]{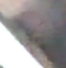} 
				& \includegraphics[width=.17\textwidth,valign=t]{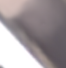} 
				\\
				& Input & 31.40dB &  33.66dB
				\\
				&   & CBDNet & DANet 
				\\
				&\includegraphics[width=.17\textwidth,valign=t]{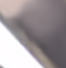} 
				&\includegraphics[width=.17\textwidth,valign=t]{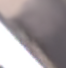}
				&\includegraphics[width=.17\textwidth,valign=t]{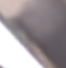}
				\\
				& 34.22dB &   34.08dB &   \textbf{34.48dB}
				\\
				Noise Image  & InvDN & VDN & Thunder (Ours)
		\end{tabular}}
	\end{center}
	\vspace{-0.2cm}
	\caption{Denoising examples from DND~\cite{dnd}. }
	\vspace{-0.1cm}
	\label{fig:dnd example}
\end{figure*}

\section{Experiments}
\subsection{Experimental Setup}
\noindent\textbf{Datasets.}
To guarantee fair comparison, Thunder is only trained on the SIDD dataset.
Both of these two dataset provide the benchmark for comparison. The results are available at the official website, the ``-'' in Table \ref{Tab:DnD_SIDD} denotes the scores are not available in the leader board.

\textit{Smartphone Image Denoising Dataset} \cite{sidd} contains 320 image pairs for training and 40 image pairs for evaluation.
These images are acquired by 5 popular smartphone cameras from 10 scenes under different lighting conditions. 

\textit{Darmstadt Noise Dataset} \cite{dnd} consists of 50 pairs of real noisy images and corresponding ground truth images.
For each pair, a reference image is captured with the low-ISO level while the noisy image is captured with higher ISO under appropriate exposure time setup. 
All the images are captured with consumer-grade cameras of different sensor sizes.


\noindent\textbf{Evaluation Metric.}
For the performance of denoising, we use the Peak Signal-to-Noise Ratio (PSNR) and Structural Similarity (SSIM) for quantitative evaluation.
For the complexity of each model, we calculate the total amount of parameters and the inference Giga Floating-point Operations Per Second (GFlops) for the corresponding model by the official published code.

\noindent\textbf{Implementation Details.}
The proposed Thunder is trained end-to-end by taking the cropped $144 \times 144$ size patches as input.
The TSE has 2 EB, and K is set to 2.
Each EB has 4 TSBs to extract the sub-bands correlation and the global features.
Similar to TES, the SPR has K down-sample times as well.
The subspace dimensionality Q in SPB  is set to 8.
Each SPB has 2 NHL blocks.
The thumbnail $\bm{T_j^i}$ channel number, as well as the channel number of high-frequent signal $\bm{S_j^i}$ feature is set to 3.
The Adam optimizer with the momentum of $\beta_1=0.9, \beta_2=0.999$ is adopted to train our network with the learning rate being initially set to 2e-4.
The learning rate will decay by half every 50k iterations by the multi-step decreasing strategy.
The training process takes 250k minibatch iterations.
The batch size is set to 96. 
The $\alpha$ and $\beta$ are set as 0.6 and 0.4, respectively.
Some common data augmentation methods are used during training, \egs horizontal flip, cropping, and rotation.
The \textbf{codes} and \textbf{more results} are provided in the supplementary material.

\subsection{Comparisons on Denoising Datasets}
The main idea for this paper is to achieve fast denoising with a lightweight network.
A fewer improve of PSNR is not always consistent with the visual quality, therefore, we think the balance of the PSNR and computational cost is more important. 
Furthermore, we think the Gflops is more important than parameters amount.

\noindent\textbf{Results on SIDD Benchmark.}
As shown in Table \ref{Tab:DnD_SIDD}, all the statistical data is computed by the official published model or copied from the corresponding articles. 
The early learning-based methods, \ies, DnCNN and FFDNet, occupy fewer parameters and GFlops due to their naive design of the network architecture but get poor performance when applied to real-world denoising tasks.
In the up part of Table \ref{Tab:DnD_SIDD}, the heavy-loaded models (Params over \textbf{10M}) can achieve better PSNR and SSIM with a huge amount of parameters and computational cost.
Thunder provides the SOTA result on the SIDD test set, and achieves 39.47dB PSNR and 0.957 for SSIM metric with 2.68M parameters and 18.81 GFlops.
It is worth noting that, Thunder outperforms the state-of-the-art InvDN and DeamNet by \textbf{0.12dB} and \textbf{0.19dB} PSNR, which is a large improvement in the field of real-world denoising.
Besides, Thunder only takes \textbf{40\%} and \textbf{12\%} Gflops of InvDN and DeamNet, respectively.
For the qualitative evaluation as shown in Fig. \ref{fig:sidd_dnd_example}, it can be concluded from the visual results that the previous methods are vulnerable to heavy noise.
However, Thunder is able to alleviate this issue through dividing the noisy residual learning schedule into two steps.

\noindent\textbf{Results on DND Benchmark.}
For each image in DND, we crop it into 20 patches with a size of $256 \times 256$ for testing.
The model submitted to the SIDD test-set is the same as that submitted to DND benchmark for fair comparison.
We can find that the heavy-loaded models all achieve impressive results due to their huge parameters. 
Because the sensors of DND is different from that of SIDD, more computational costs represent robustness and generalization of the model.
As can be demonstrated from quantitative results shown in Table \ref{Tab:DnD_SIDD}, among lightweight models, although the highest results (39.63dB and 0.955) is obtained by DeamNet, Thunder finally gets 39.57dB PSNR and 0.9526 SSIM by taking about one tenth Gflops (146.36 vs 18.81).
In addition, the results of DeamNet are trained both on the RENOIR and SIDD, while Thunder only trained on SIDD.
For the qualitative comparison as in Fig. \ref{fig:dnd example}, the signal-reconstruct-based Thunder can avoid the over-smoothing problem by accurately preserving the high-frequency details.

\begin{table}[tb]
	\centering 
	\small
	\begin{tabular}{c|c|c|c}
		\toprule
		\tabincell{l}{Model} & PSNR (dB)$\uparrow$  & Params(M) $\downarrow$ & Gflops(G) $\downarrow$\\
		\midrule 
		\tabincell{l}{Thunder} & \textbf{39.50}  & 2.68 & 18.81 \\
		\tabincell{l}{w/o SPR} & 38.49 &2.19 	& 14.57  \\
		\tabincell{l}{w/o Projection} & 39.43 & 2.77 &	19.68\\
		\tabincell{l}{w/o CSP} & 39.33 &2.68 & 18.81 \\
		\tabincell{l}{w/o Global} & 39.34 & \textbf{1.32} & \textbf{10.08} \\
		\tabincell{l}{NHL $\rightarrow$ Inv} & 39.38 & 3.36 & 30.0 \\
		\bottomrule
	\end{tabular}
	\caption{Ablation study on SIDD validation dataset.}
	
	\label{Tab:Ablation_s}
	\vspace{-0.2cm}
\end{table}

\subsection{Ablation Study}

\noindent\textbf{Effect of SPR.}
The SPR is designed as the signal refiner in Thunder.
Based on the coarse results $\bm{I_t}$ obtained by the TSE, SPR first decomposes the input into low and high frequency groups, then refines the high-frequent features for saving the computational costs.
To show the importance of SPR, we discard the SPR module and only use the  $\bm{I_t}$ as final output.
As shown in Table \ref{Tab:Ablation_s}, the PSNR drops significantly to 38.49dB (Denoted as w/o SPR).

\begin{figure}[t]
    \centering
    \scriptsize
    \begin{minipage}[b]{0.54\textwidth}
        \centering
        \includegraphics[width=\linewidth]{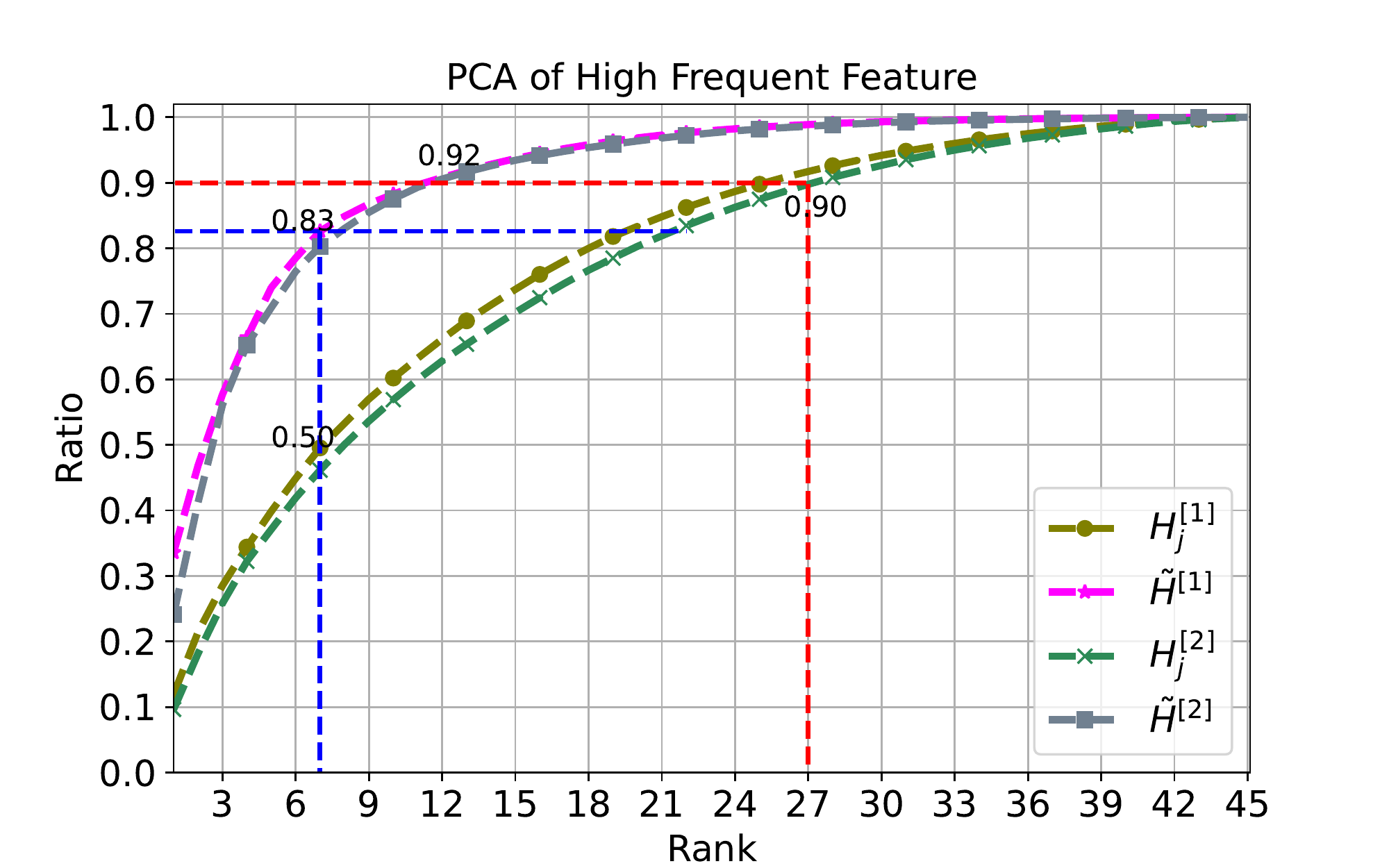}
    \end{minipage}
    \hfill
    \begin{minipage}[b]{0.44\textwidth}
        \centering
        \includegraphics[width=\linewidth]{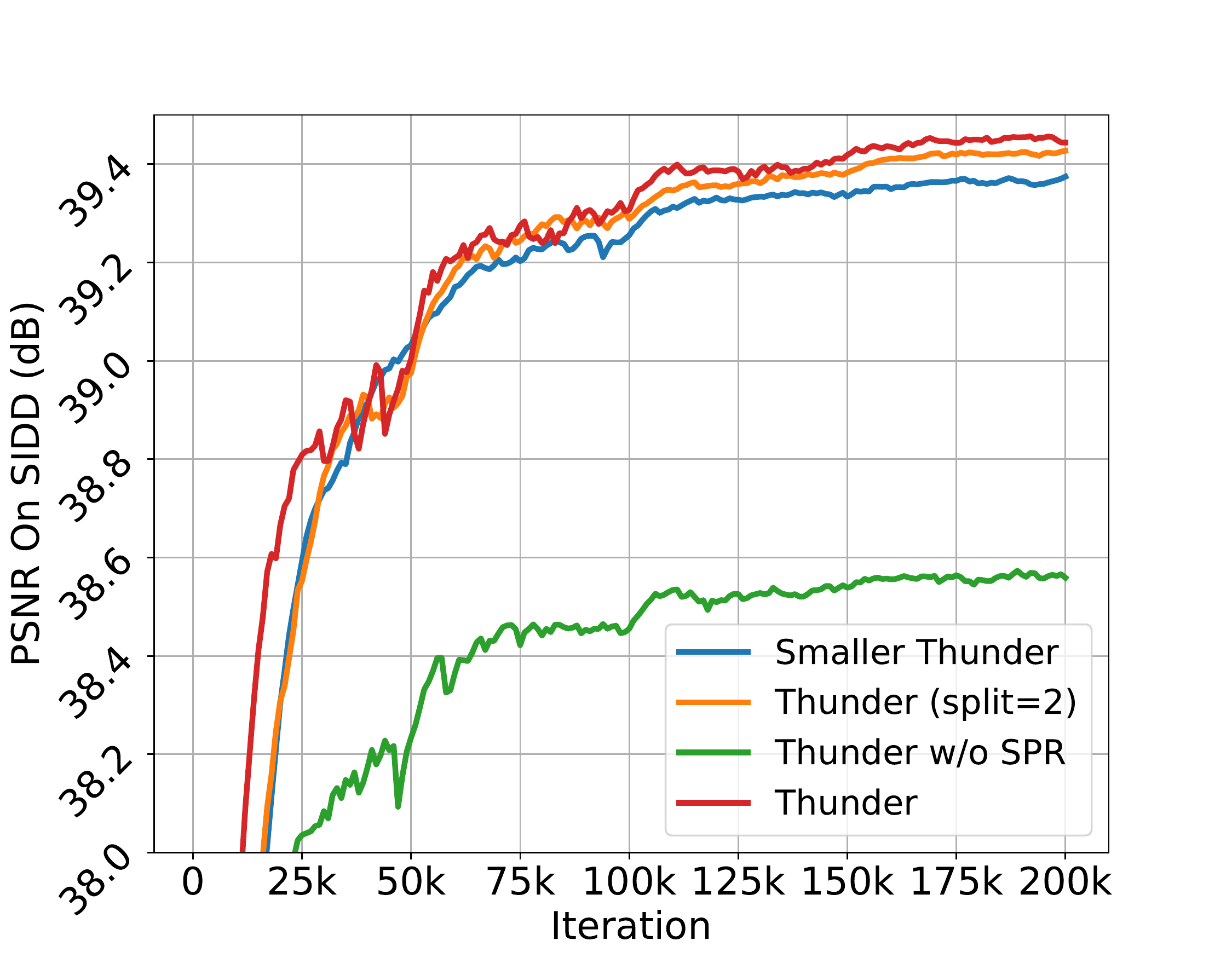}
        
    \end{minipage}
    \caption{
        (Left) The PCA of $H_j$ and $\tilde{H}$. The feature maps are sampled from two different cases which distinguished by the superscripts. The rank denotes the PCA components number. $\tilde{H}$ can leverage fewer principle components to represent the feature.
        (Right) Validation PSNR curve for the baselines of Thunder.
    }
    \label{fig:lr_and_psnr_curve}
    \vspace{-0.1cm}
\end{figure}

	

\begin{figure}[t]
	\begin{center}

		\vspace{-2mm}
		\small
		\begin{tabular}[t]{c@{}c@{}c@{}c@{}c@{}}
			\hspace{-8mm}
			\includegraphics[width=.25\linewidth]{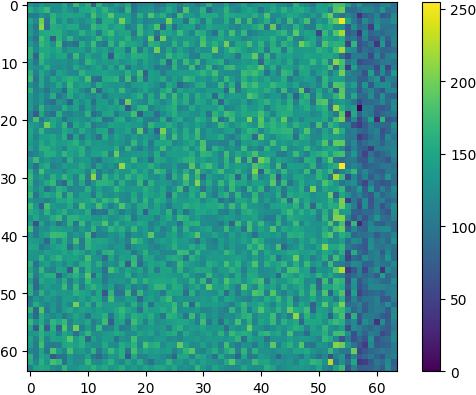}& 
			\includegraphics[width=.25\linewidth]{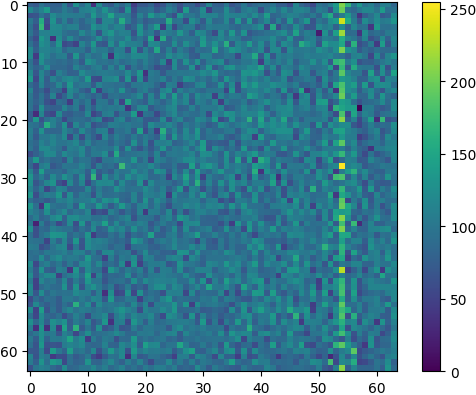}& 
			\includegraphics[width=.25\linewidth]{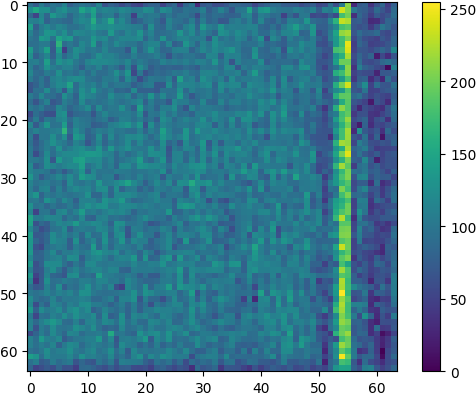}& 
			\includegraphics[width=.25\linewidth]{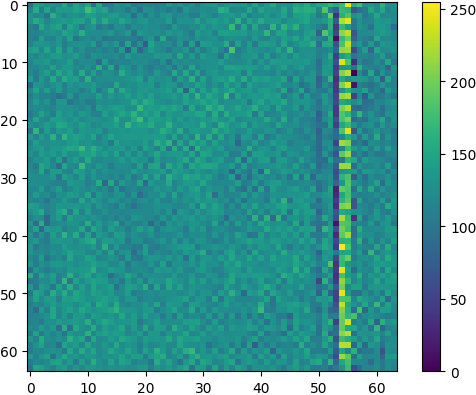}& 
			\hspace{-4mm}
			\\
			\hspace{-8mm}Before GA &After GA  &  Before NHL & After NHL 			
			
		\end{tabular}
	\end{center}
\vspace{-7mm}	
	\caption{ Average feature map visualization of GA and last NHL.}
	\label{fig:feats_r2}
	\vspace{-0.5cm}
\end{figure}

\noindent\textbf{Effect of Projection Matrix.}
The subspace projection module tranforms the $\bm{H_j}$ which contains the noise and the high-frequent details to the signal subspace.
Considering the fact that the signal lies in the low-rank subspace, the projected feature is more likely to preserve the signal information and reduce the noise.
In order to show the effect of the projection computation, we experiment SPR without the subspace projection (Denoted as w/o Projection).
Specifically, instead of projecting $\bm{H_i}$, we use some Res blocks to compute the attention $\bm{A_H}$ based on $\bm{\hat{H}}$ and get the $\tilde{\bm{H}}$ according to $\bm{A_H} \times \bm{H_i} + \bm{\hat{H}}$, instead of Eq. \ref{SP}.
Although this design takes more parameters and Gflops, it achieves lower PSNR, which demonstrates that the projection matrix can help our model to learn the signal features.
In order to verify the effect of the projection matrix in decreasing the complexity of the feature space, we use the PCA\cite{pca} to analyze the high frequent feature before and after the projection as shown in Fig. \ref{fig:lr_and_psnr_curve}.

\noindent\textbf{Effect of CSP module.}
In the TSE, the spatial scale of the output RGB thumbnail is not the same as the original input results in the mismatch problem, therefore, the CSP subnet is proposed to project the thumbnail $\bm{T}$ to $\bm{I_t}$.
This procedure can be viewed as an interpolation (from size $\frac{1}{K} \rightarrow 1$), however, simply performing the linear interpolate function such as bicubic interpolation results in lower PSNR as shown in Table \ref{Tab:Ablation_s} (Denoted as w/o CSP).

\noindent\textbf{Effect of Global Information.}
The global branch in TSE can enrich the global features of sub-bands.
In order to study the effect of $\bm{G}_j^i$, we transform the TSB to one stream manner.
The GA module is also discarded.
As demonstrated in Table \ref{Tab:Ablation_s} (w/o Global), the two-stream design can improve the performance of Thunder with a little computational cost and parameters. 
As shown in Fig.\ref{fig:feats_r2}, the GA module prefers to modify global information of corresponding features where the whole feature value is suppressed in this case.
As for NHL module, some pixel-wise and global modifies can be found here, which demonstrates NHL is able to process the global and local information.
The global information can help to improve the performance from 39.34 to 39.50 (Table \ref{Tab:Ablation_s}).

\noindent\textbf{The Effect of Sub-bands Correlation.}
The proposed NHL can jointly leverage the sub-bands mutual information between $\bm{N}$, $\bm{H}$ and $\bm{T}$.
The correlation between the sub-bands extracted by the HWT plays an important role in boosting the denoising performance.
To illustrate the effect of NHL, we also experiment without it.
Firstly, instead grouping the $\bm{HT_j^i}$ into $\bm{N}$, $\bm{H}$ and $\bm{T}$, we study the Thunder which decomposes $\bm{HT_j^i}$ into two groups, the PSNR will drop to 39.42dB (Fig. \ref{fig:lr_and_psnr_curve}: Thunder split=2).
Also, we feed $\bm{HT_j^i}$ to some invertible residual-based blocks (similar with \cite{invdn}) and the PSNR drops from 39.5dB to 39.38dB  (Table \ref{Tab:Ablation_s}: NHL $\rightarrow$ Inv).

			

\subsection{Further Analysis}
\begin{figure}[t]
    \centering
    \scriptsize
    \begin{minipage}[t]{0.44\textwidth}
        \centering
        \begin{tabular}{c@{}c@{}c@{}}			
			\includegraphics[width=.41\linewidth]{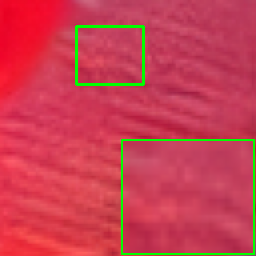}& \hspace{2mm}
			\includegraphics[width=.41\linewidth]{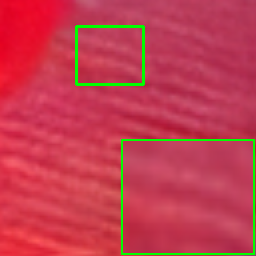}& 
			\\
			(a) w $\mathcal{L}_t$ & \hspace{6mm}(b) w/o $\mathcal{L}_t$  & 
			
		\end{tabular}
		\caption{Thumbnail for Thunder with and w/o $\mathcal{L}_t$.}
		\label{fig:noLt_thumbnails}
    \end{minipage}
    \hfill
    \begin{minipage}[t]{0.54\textwidth}
        \centering
        \begin{tabular}{c|c|c|c}
		\toprule
		\tabincell{l}{Model} & PSNR (dB)$\uparrow$  & Para.(M)$\downarrow$ & Gflo.(G) $\downarrow$\\
		\midrule 
		\tabincell{l}{Thunder} & \textbf{39.50}  & 2.68 & 18.81 \\
		\tabincell{l}{1/2 Thumbnail} & 39.14 & \textbf{0.597} &\textbf{9.98} \\
		\tabincell{l}{1/8 Thumbnail} & 39.49 & 17.26 &31.23 \\
		\tabincell{l}{Noisy Residual} & 39.01  & 2.68  & 18.81 \\
		\tabincell{l}{Squ-DS} & 39.48 & 2.68 & 18.81  \\
		\tabincell{l}{Conv-DS} & 39.46 & 2.71 	& 19.0  \\
		\tabincell{l}{DB2 Wavelet} & 39.47 & 2.68 & 18.81 \\
		\tabincell{l}{Smaller Thunder} & 39.39 & 1.62 &11.68  \\
 		
		\bottomrule
	\end{tabular}
    \captionof{table}{Further analysis for the design of Thunder.}
    \label{Tab:Further_Analysts}
    \end{minipage}

    \vspace{-0.4cm}
\end{figure}

\begin{figure}[!htb]
	\centering
	\scalebox{1}{
	\begin{tabular}[t]{c@{ }c@{ } c@{ } c}
		\includegraphics[width=.22\linewidth]{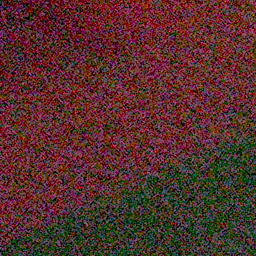}&  
		\includegraphics[width=.22\linewidth]{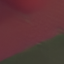}&  
		\includegraphics[width=.22\linewidth]{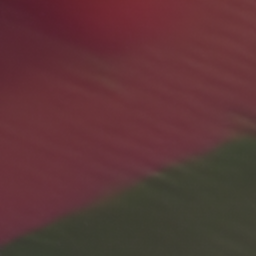}&  
		\includegraphics[width=.22\linewidth]{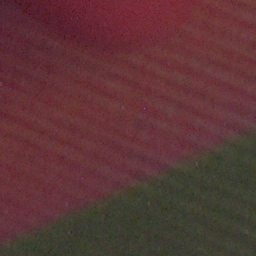}
		\\
		
		\includegraphics[width=.22\linewidth]{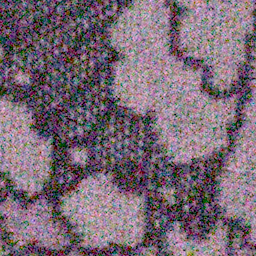}& 
		\includegraphics[width=.22\linewidth]{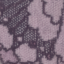}&   
		\includegraphics[width=.22\linewidth]{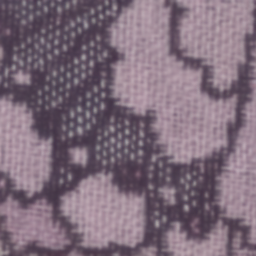}& 
		\includegraphics[width=.22\linewidth]{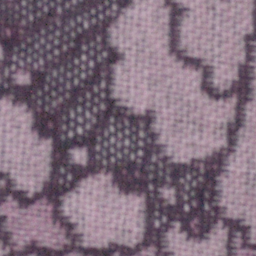} 
		\\
		Noisy  &  Thumbnail &  Thunder-Ours & GT\\	
	\end{tabular}
	}
	\caption{Visualization of Thumbnail (Up-sampled to original size) and final output by SPR. SPR is able to enrich the high-frequent details of thumbnail.}
	\vspace{-0.4cm}
	\label{fig:thumbnail}
\end{figure}

\noindent\textbf{Is the RGB Thumbnail Subspace Important?}
The loss $\mathcal{L}_t$ for $\bm{T}$ can restrict Thunder to project the feature to the RGB thumbnail subspace.
Without this loss, the network is more likely to learn the projection subspace by itself.
To demonstrate the importance of RGB thumbnail subspace, we train the Thunder without $\mathcal{L}_t$ on $\bm{T}$, thus the network can learn the projection subspace by itself.
It is worth noting that, although the learned subspace is not the RGB thumbnail subspace, the CSP can project this subspace to clean subspace and obtain $\bm{I_t}$, and finally refines it based on SPR.
However, the performance of Thunder under that setting will drop to 39.47dB which denotes that, the loss $\mathcal{L}_t$ can boost the model and improve the denoising capability.
As shown in Fig.\ref{fig:noLt_thumbnails}, the thumbnails of Thunder under supervised loss $\mathcal{L}_t$ are worse than without it, especially for some textural details, but finally achieve higher PSNR.
It indicates that the loss $\mathcal{L}_t$ can help TSE to disentangle the low and high frequent sub-bands, and a better disentanglement can finally achieve higher PSNR.
We also show some visualizations for $\bm{T}$ in Fig. \ref{fig:thumbnail}, the thumbnails suffer from blur artifacts, and proposed SPR can enrich high-frequent details for the thumbnails.

	

\noindent\textbf{What is the Appropriate Size of Thumbnail?}
The size of the thumbnail should be carefully set to guarantee the trade-off between signal preservation and noise reduction.
The bigger thumbnail contains more information on both signal and the noise which results in poor PSNR.
When setting the thumbnail smaller by considering the capability limitation, the denoising results will degrade seriously.
Results are shown in Table \ref{Tab:Further_Analysts}, the 1/2, 1/8 denotes to use the $\frac{1}{2}H \times \frac{1}{2}W$ and $\frac{1}{8}H \times \frac{1}{8}W$ size thumbnail, respectively.
The 1/8 Thumbnail needs more downsample and upsample layers to achieve the 1/8 size thumbnail, calling for more network parameters.

\noindent\textbf{End-to-end Residual Learning vs Thunder.}
When the end-to-end residual learning approaches estimate the noise residual, it implicitly estimates the signal, which doubles the computational burden.
Therefore these methods demand more channel numbers (filters numbers) to represent the noisy feature.
As mentioned before, we depart the end-to-end residual learning into two stages.
In order to illustrate this, we transform the Thunder to end-to-end noisy-residual learning manner by skipping adding the final output to the input.
Shown in Table \ref{Tab:Further_Analysts}, the PSNR drops from 39.5dB to 39.01dB.

\noindent\textbf{About the Haar Wavelet Transform.}
We follow the existing work \cite{invdn,MWCNN} to use the Haar wavelet transform because of their simplicity and provided efficiency.
We test the performance of Thunder with different downsample operations such as the convolution-based downsample layer (Con-DS) and squeeze-based downsample layer (Squ-DS).
The Con-DS occupies more parameters and GFlops and gets lower PSNR of 39.46dB.
Instead of using the Haar Transform, we conduct the experiment by using the other high-order Daubechies wavelet (DB2). 
The results in Table \ref{Tab:Further_Analysts} demonstrated that using Haar wavelet for downsampling and upsampling in network is the best choice in terms of quantitative evaluation.

\noindent\textbf{Can Thunder be Smaller?}
We study the denoising capability of Thunder with different numbers of TSB in each EB.
Smaller Thunder is composed by \textbf{M}=2 TSBs.
The small Thunder gets 39.39dB PSNR with 1.62M parameters and 11.68 GFlops, however the performance of PSNR drops from 39.5dB to 39.39dB.
We find that \textbf{M}=4 is appropriate to balance the capability and computational cost.



\section{Conclusion}
In this paper, we propose Thunder that jointly leverages Thumbnail Subspace Encoder (TSE) and Subspace Projection-based Refine Module (SPR) to achieve fast real-world denoising.
We divide the end-to-end residual learning mechanism into two steps.
To accelerate the denoising, we compressed the complexity of the projection subspace to the Thumbnail subspace.
To preserve the capability of denoising, the subspace projection method is exploited to reduce the noise and also preserve the signal details in high-frequent features, thus refining the thumbnail.
Extensive experiments demonstrate that Thunder consuming fewer computational costs is able to achieve competitive results.

\clearpage
%
%
\bibliographystyle{splncs04}
\bibliography{egbib}
\end{document}